\documentclass[runningheads]{llncs}
\usepackage{graphicx}
\usepackage{amsmath,amssymb} % define this before the line numbering.
\usepackage{lineno}
\usepackage{color}
\usepackage{times}
\usepackage{epsfig}
\usepackage{wrapfig}
\usepackage{color}

\newcommand{\omitme}[1]{}

\begin{document}
%\renewcommand\thelinenumber{\color[rgb]{0.2,0.5,0.8}\normalfont\sffamily\scriptsize\arabic{linenumber}\color[rgb]{0,0,0}}
%\renewcommand\makeLineNumber {\hss\thelinenumber\ \hspace{6mm} \rlap{\hskip\textwidth\ \hspace{6.5mm}\thelinenumber}}
%\linenumbers
\pagestyle{headings}
\mainmatter
\def\ECCV10SubNumber{1289}  % Insert your submission number here

\title{Optimization of Weighted Curvature for Image Segmentation} % Replace with your title

%\titlerunning{ECCV-10 submission ID \ECCV10SubNumber}

%\authorrunning{Noha El-Zehiry and Leo Grady}

\author{Noha El-Zehiry and Leo Grady}
\institute{Department of Image Analytics and Informatics \\Siemens Corporate Research\\ Princeton, NJ}

\maketitle

\begin{abstract}
Minimization of boundary curvature is a classic regularization
technique for image segmentation in the presence of noisy image
data. Techniques for minimizing curvature have historically been
derived from descent methods which could be trapped in a local minimum
and therefore required a good initialization.  Recently, combinatorial
optimization techniques have been applied to the optimization of
curvature which provide a solution that achieves nearly a global
optimum.  However, when applied to image segmentation these methods
required a meaningful data term. Unfortunately, for many images,
particularly medical images, it is difficult to find a meaningful data
term.  Therefore, we propose to remove the data term completely and
instead \emph{weight} the curvature locally, while still achieving a
global optimum.
\end{abstract}

\section{Introduction}

A classic prior model for the boundary of objects in an image is that
the boundary have a small curvature.  This model was proposed and
theoretically justified by Mumford \cite{MumfordComputerVision} and
also appeared in the first active contour work by Kass, Witkin and
Terzopoulos \cite{179} who proposed an optimization of the boundary
curvature.  Subsequently, the optimization of boundary curvature
became a common feature of variational methods for active contours and
level sets
\cite{caselles1997geodesic,ChanInpainting,EsedogluInpainting}.
However, all of these methods use descent-based optimization, causing
the solution to get stuck in a local minimum and depend strongly on
having a good initialization.  Following the description by Bruckstein
{\it{et al.}}  \cite{EpiConvergence} of the curvature of a polygon,
the curvature of an object boundary was formulated on a graph by
Schoenneman {\it{et al.}} \cite{CremersCurvature} (and previously in a
different manner by the same authors \cite{ElasticRatio}).
Specifically, on the graph \emph{dual} to the pixel lattice, the
boundary of an object may be described by a polygon comprised of graph
edges, and therefore the boundary having minimum curvature could be
found by optimizing over all polygons which had a curvature value as
defined by Bruckstein {\it{et al.}}.  However, this optimization
required a long computation time (minutes to hours) and often did not
find a solution achieving a global optimum.  Therefore, this work was
followed by El-Zehiry and Grady \cite{CurvatureCVPR10} who used a \emph{primal}
formulation of the curvature energy that parameterized the boundary
polygon in terms of the normal vectors and computed a solution in
seconds which often achieved a global optimum.  For simplicity, in
this work we consider only the segmentation of an object from a
background, i.e., a two-class image segmentation problem.

All of these previous approaches to utilizing curvature minimization
employed curvature as a boundary \emph{regularization} for a data term
which modeled the object and background intensity/color/texture/etc.
Since this data term was imperfect (due to noise), the curvature
regularization was used to regularize the solution.  However, in
practice, this approach can suffer from two problems: 1) It may be
hard to model the object and background data {\it{a priori}} or,
worse, the distributions can significantly overlap (or even be equal),
2) Sometimes the true object boundary has a high curvature, but a
minimum curvature regularization can cause the boundary to become
erroneously smooth (e.g., by cutting of corners).  In particular, the
data modeling problem often occurs in medical imaging.  For example, a
tumor may be made of the \emph{same} tissue as the organ it is
attached to, such that its intensity distribution appears identical to the organ's intensity in
the acquired image (meaning that only spatial information can be used
to distinguish tumor from healthy tissue).  Another example is in the
segmentation of heart chambers in which each chamber is filled with
blood (i.e., has the same appearance) and, worse, an open
valve between the chambers means that only the spatial information is
capable of distinguishing one chamber from another. Consequently, in
this paper we highlight the ability of our method to operate in these
difficult circumstances encountered in medical imaging, with the
understanding that the technique could be applied to any arbitrary
images encountered in computer vision.

One method which has been used in past to address the unreliability of
a data term is to simply employ contrast-dependent \emph{weights} on
the graph edges, which requires the optimization of only a single
term.  However, to avoid a trivial solution, previous methods have
employed \emph{seeds}; a foreground seed is a small subset of pixels that have been labeled as belonging to object and a background seed is a small subset of pixels that
have been labeled as belonging to background.  These seeds may be
obtained interactively from a user who is specifying a particular
object (e.g.,
\cite{boykov2001interactive,rother2004grabcut,grady2005:random}) or
automatically from a system trained to look for a particular object
(e.g., \cite{funka2006automatic,wighton2009fully}).  For example, this
type of approach which uses seeds and contrast-sensitive edge
weighting has been employed with such popular algorithms as graph cuts
\cite{boykov2001interactive}, random walker \cite{grady2006random},
geodesic segmentation \cite{bai2007geodesic} and power watersheds
\cite{couprie2009power}.  We propose to apply this same approach in
the curvature method of \cite{CurvatureCVPR10}, which has the effect
of allowing object boundary curvature to be high when supported by
image contrast, but penalizing boundary curvature for weak or noisy
object boundaries. In this way, our formulation solves both of the
problems outlined above that can be encountered when a curvature term
is used as regularization for a data model, since the formulation does
not require a reliable data prior and also permits high-curvature
boundaries when supported by data.

\omitme{\textbf{Maybe we should remove this as we have illustrated two or three examples with sharp corners}
Although we focus in this paper on
image segmentation problems which are challenging due to unreliable
data terms, note that weighted curvature could also be used in
conjunction with a data term to permit high-curvature objects which
are supported by image data.}

The use of a weighted curvature objective function is designed to
overcome problems with unreliable data terms and permit a
high-curvature object boundary when supported by image contrast.
Unfortunately, the authors of \cite{CurvatureCVPR10} perform the
optimization of object curvature on a graph using the Quadratic
Pseudo-Boolean Optimization with Probing (QPBOP) method, which can be
expected to produce \emph{all} unlabeled pixels when there is no unary
(data) term \cite{MinimizingNonSubmodular,RoofDuality}.  Therefore,
QPBOP will not allow us to find a global optimum of the weighted
curvature functional when no data term is present. However, we show
that by introducing an \emph{attraction force}  \omitme{to promote the addition
of object-labeled pixels (assuming that the size of the object is
smaller than the background)}, we can decrease the number of negative weighted edges in the formulation presented in \cite{CurvatureCVPR10} which allows QPBOP to find the optimal solution.

After describing our image segmentation technique, we demonstrate that
it can be applied to control images that were designed to have
\emph{zero} difference between the intensity distribution of the
object and background, as well as having large sections of missing
boundary information.  Although other techniques are known for
robustness to weak boundaries, such as graph cuts and random walker,
we show that they are unable to address these difficult cases, while
our algorithm can.  Older active contour methods could also be applied
to solve some of these image segmentation problems, but they rely on a
good initialization and a setting of parameters to tradeoff between
multiple terms which often needed to be set individually for each
image.  Following these control images, our algorithm is applied to a
series of medical image segmentation problems that exhibit the same
challenges which were observed in the control set.

\section{Methods}
\vspace{-0.1in}
We begin this section with a review of the 2D curvature optimization
framework presented in \cite{CurvatureCVPR10} before proceeding to our contrast weighted curvature formulation, optimization and addition of an attraction force.
\subsection{Curvature Energy}

The continuous formulation of Mumford's Elastica model is
defined for curve $\mathcal{C}$ as
\begin{equation}
E(\mathcal{C})= \int_{\mathcal{C}} (a+ b\kappa^2) ds \quad  a,b>0
\end{equation}
 where $\kappa$ denotes the scalar curvature and $ds$ represents the
arc length element. When $a=0$ (the arc length is ignored), the model
reduces to the integral of the boundary squared curvature
$E(\mathcal{C})=\int_{\mathcal{C}} \kappa^2 ds$.

The use of combinatorial optimization by \cite{CurvatureCVPR10} to minimize the elastica model prompted the discrete formulation of the curvature on a
graph. A graph $\mathcal{G}=\{ \mathcal{V}, \mathcal{E}\}$ consists of
a set of vertices $ v \in \mathcal{V}$ and a set of edges $e \in
\mathcal{E} \subseteq \mathcal{V} \times \mathcal{V}$. An edge
incident to vertices $v_i$ and $v_j$ is denoted $e_{ij}$. In our
formulation, each pixel is identified with a node, $v_i$. A
weighted graph is a graph in which every edge $e_{ij}$ is assigned a
weight $w_{ij}$. An edge cut is any collection of edges that separates
the graph into two sets, $\mathcal{S} \subseteq \mathcal{V}$ and
$\overline{\mathcal{S}}$, which may be represented by a binary
indicator vector $x$,
\begin{equation}
x_i = \begin{cases}
  1 \quad \text{if $v_i \in \mathcal{S}$},\\
  0 \quad \text{else}.
\end{cases}
\end{equation}
The cost of the cut represented by any $x$ is given by
\vspace{-0.05in}
\begin{equation}
\text{Cut}(x) = \sum_{e_{ij}} w_{ij} |x_i - x_j|.
\label{eq:cut}
\vspace{-0.02in}
\end{equation}

Bruckstein \textit{et al.} \cite{EpiConvergence} expressed the
curvature of a 2D polygon in terms of the angular change
 between consecutive polygonal segments.

%\begin{figure}[h!]
%\vspace{-0.5in}
%\begin{center}
%\begin{tabular}{ccc}
%%\fbox{\rule{0pt}{2in} \rule{0.9\linewidth}{0pt}}
%   \includegraphics[width=0.35\linewidth]{./Figures_PDF/BoundaryVSExterior2.pdf}&
%   \includegraphics[width=0.3\linewidth]{./Figures_PDF/6PointNeighborhood.pdf}&
%   \includegraphics[width=0.3\linewidth]{./Figures_PDF/AttractionConstruction_1.pdf}\\
%    (a) &  (b) & (c)\\
%   \end{tabular}
%   \end{center}
%  \caption{(a)The exterior angle $\phi$ formed by the adjacent line
%  segments in a polygonal curve and the interior angle $\alpha$ formed
%  by the intersection of the normals to the line segments, $\phi$ =
%  $\alpha$. (b) Six point neighborhood system used in evaluating the 3D curvature. (c) The corresponding construction to the attraction force $x_1 x_2$.}
%\label{Neighborhood}
%\end{figure}

Instead of a polygon, it was observed in \cite{CremersCurvature} that the
polygon could be viewed as existing on a \emph{dual} graph.  However,
in \cite{CurvatureCVPR10} the same idea was formulated on a \emph{primal} graph
in order to permit an easier optimization and more generalizable
formulation.  In this formulation, if two edges $e_{ij}$ and $e_{ik}$, incident to a node $v_i$, are cut then the cut is penalized with
value $w_{ijk}=\frac{\alpha^p}{min(||\overrightarrow{e_{ij}}||,
||\overrightarrow{e_ik}||)}$, where $\alpha$ is the angle between the
edges.  This cut penalty was then exactly decomposed into three edge
weights
\begin{equation}
E(x_i,x_j,x_k)=w_{ij}|x_i-x_j| + w_{ik}|x_i-x_k| - w_{jk}|x_j-x_k|,
\label{DiscreteCurvature}
\end{equation}
where $w_{ij} = w_{ik}=w_{jk}= \frac{1}{2} w_{ijk}$. Therefore, the
minimum cut with respect to these edge weights is a cut that minimizes
Bruckstein's discrete curvature formulation on the dual graph.
Despite the negative weights, it was shown in \cite{CurvatureCVPR10} that QBPOP
was able to find a minimum cut in most circumstances. Notice that
although the curvature clique was designed to penalize the cut of both
edges $e_{ij}$ and $e_{ik}$, the decomposition to pairwise
interactions add an edge $e_{jk}$ with negative weight. We denote the
the set of effective edges with nonzero weights as $\mathcal{E}^*
\supseteq \mathcal{E}$.

\subsection{Weighted Curvature}
\label{sec:weightedCurvature}

To perform curvature based segmentation in \cite{CremersCurvature,CurvatureCVPR10}, data term had to be added to the curvature model to obtain the boundary of the object of interest. To solve the segmentation problem without dependence on a data term, we propose to weight the curvature term locally based on the image intensity profile.
The primal pairwise curvature formulation in \cite{CurvatureCVPR10} makes it feasible to weight the curvature based on the intensity differences between the terminal points of a particular edge.\\

\begin{figure}
\vspace{-2.4in}
\begin{center}
  \includegraphics[width=0.9\linewidth]{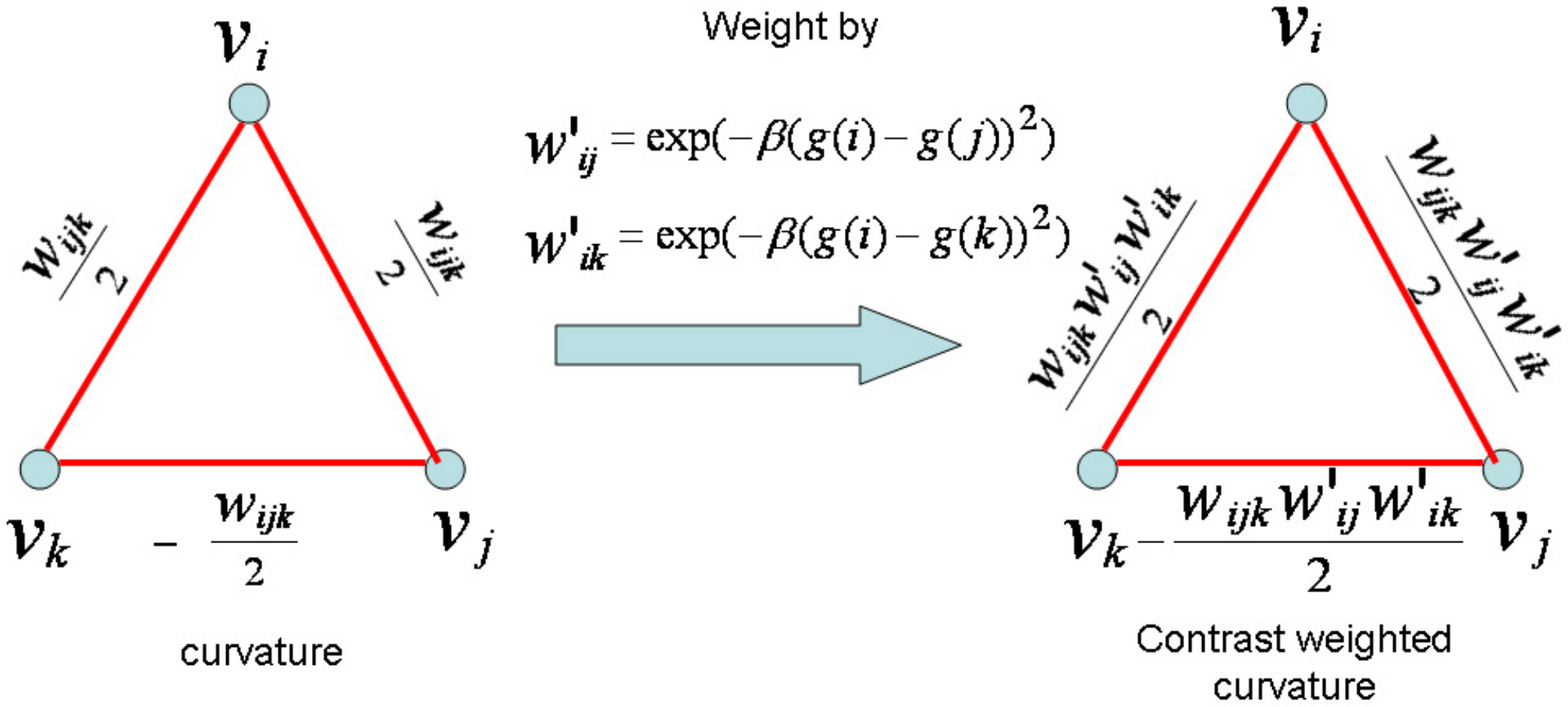}
  \vspace{-2.1in}
  \caption{Weighted curvature clique versus unweighted curvature clique.}
    \end{center}
    \vspace{-0.3in}
    \label{Fig:construction}
\end{figure}

According to  the curvature formulation in \cite{CurvatureCVPR10},  edges $e_{ij}$ and $e_{ik}$ are cut when the pixel $i$ is a foreground pixel and $j$ and $k$ are background pixels or vice versa. Therefore, the curvature clique formed by these edges should be weighted by the intensity differences between pixels $i$ and $j$ and the intensity difference between $i$ and $k$. \\
This can be formulated as follows: Given a 2D image with image values associated with each pixel (node),
$g: \mathcal{V} \rightarrow R$. The weighted curvature $wc_{ijk}$ is given by

\begin{equation}
wc_{ijk}=w_{ijk}\; w'_{ij} \; w'_{ik},
\end{equation}
where
\vspace{-0.2in}
\begin{eqnarray}
w'_{ij}&=& exp(-\beta(g(i)-g(j))^2),\\
w'_{ik}&=& exp(-\beta(g(i)-g(k))^2).
\end{eqnarray}

 \noindent The parameter $\beta \geq 0$ controls the contract strength. The cut penalty is calculated using the same decomposition in \ref{DiscreteCurvature} with weights $w_{ij}$=$w_{ik}$=$w_{jk}$=$\frac{wc_{ijk}}{2}$ depicted in Figure 1.

 \omitme{And the weighted curvature energy is represented by
   \begin{equation}
E(x_i,x_j,x_k)=wc_{ij}|x_i-x_j| + wc_{ik}|x_i-x_k| - wc_{jk}|x_j-x_k|,
\label{DiscreteCurvature}
\end{equation}
where $wc_{ij} = wc_{ik}=wc_{jk}= \frac{1}{2} wc_{ijk}$}

The contrast weighted curvature regularization eliminates the dependance on the data model and  provides better segmentation of high curvature features in the image (such as sharp corners) when supported by high contrast.

\subsection{Optimization}
\label{sec:optimization}

In the previous sections, the original Bruckstein formulation of
polygonal curvature was transformed into the problem of finding a
minimum cut on a graph in which some of the edge weights were
negative.  Unfortunately, the negative edge weights introduced by the
third term of \eqref{DiscreteCurvature} causes the minimum cut problem
to be nonsubmodular \cite{WhatEnergy}, \textit{i.e.}, straightforward
max-flow/min-cut algorithms will not yield a minimum cut.  However, it
was shown in \cite{CurvatureCVPR10} that the Quadratic Pseudo Boolean
Optimization (QPBO) \cite{MinimizingNonSubmodular} and Quadratic
Pseudo Boolean Optimization with Probing (QPBOP)\cite{RoofDuality}
offered a solution to the optimization problem that frequently offered
a complete, optimal solution.

A practical problem that we encountered due to the elimination of the data model (unary term or terminal links in the graph) is that the optimization using QPBO/ QPBOP (used in \cite{CurvatureCVPR10}) fails to provide complete labeling. Hence, in the next section present the \emph{attraction energy} that enables the QPBO/ QPBOP to provide complete labeling.

 \subsection{Attraction Energy}

 The second contribution of this paper is the introduction of the
attraction force to the segmentation problem. The attraction force is
inspired by the intermolecular forces that maintain the structure of
the molecule and prevent it from decomposing into its salient
atoms. This scenario is very analogous to the segmentation problem
where the object to be segmented is similar to a molecule that we
would like to maintain its atoms (the pixels that constitute the
object of interest) tightly connected to each other.  Coulomb's Law
states that the force of attraction between two objects is equal to
the product of their charges $q_1$ and $q_2$ is given by
\begin{equation}
F= \frac{k}{r^2} q_1 q_2,
\label{ColumbsLaw}
\end{equation}
where $k$ is the Coulomb's constant and $r$ is the distance between
$q_1$ and $q_2$. Hence in our discrete formulation, we need to
maximize the attraction force between two variables $x_1$ and $x_2$ given
by
\begin{equation}
E_{\rm{attraction}}= \frac{w_{12}}{2}(x_1 x_2 + (1-x_1)(1-x_2)\footnote{Note that the addition of the second term here is very natural and still analogeous to \eqref{ColumbsLaw} in the sense that we have two molecules, one for the foreground and one for the background and hence we have to add two terms for the attraction force; one penalizing both pixels to be labeled 1 and another to penalize both pixels being labeled 0 }),
\end{equation}

\noindent where $w_{12}$ is the weighted curvature weight of the edge $e_{12}$ described in the previous section. Notice that maximizing this attraction force is equivalent to minimizing its negative.
%Maximizing this attraction force in the image segmentation context
%plays an analogous role to maximizing intraclass correlation in
%clustering problems.  \comment{LJG: I'm confused --- We have $w_{12}$
%here, but below it seems like $\mu$ is used instead.}
%%\textbf{Now that I am looking at Coulomb's law, may be normalizing by
%%the distance give better results, I should try this !!!}}.

A key value of the attraction force is that it allows for an
optimization of the curvature energy in the absence of  the unary term. Specifically, the construction in \cite{WhatEnergy} represents the negative attraction energy by adding edges $e_{12}$with weight $w_{12}$
, $e_{\mathcal{S}1}$ an edge $e_{2\mathcal{T}}$ with weights $\frac{w_{12}}{2}$. The addition of positive
weights changes the sign of some of the negative weights introduced by
the curvature term. This justifies our choice for the attraction weight as $\frac{w_{12}}{2}$ which makes the the weight of $e_{12}$ nonnegative. These sign changes affect the optimization problem by strongly decreasing the number of unlabeled variables in the output of the QPBOP.

\subsection{Summary}

The segmentation problem is modeled as the solution, $x$, which
minimizes the energy
\begin{equation}
 E(x) =   E_{\rm{curvature}}(x) - \lambda \; E_{\rm{attraction}}(x),
\end{equation}
for strictly positive weighting parameter $\lambda$ that controls the relative importance of the attraction energy with respect to the curvature energy.

The curvature term is written as a summation by
\begin{equation}
E_{\rm{curvature}}(x) = \sum_{e_{ij}\in \mathcal{E}^*} w_{ij}
|x_i - x_j|,
\end{equation}
and the attraction force is given by
\begin{equation}
E_{\rm{attraction}}(x) = \sum_{e_{ij}\in \mathcal{E}^*} \frac{w_{ij}}{2}(x_i x_j+(1-x_i)(1-x_j)).
\end{equation}

Foreground and background seeds are used to initialize the segmentation, a foreground seed
$v_i$ is set to $x_i = 1$ while a background seed would be set to
$x_i=0$.

\section{Experimental Results}

This section presents a sample of our segmentation results using weighted curvature and compares our segmentation to the corresponding results obtained by graph cuts and random walker.  In all of our results, we have used the 8-point connected lattice. We begin by demonstrating the usefulness of weighted curvature on synthetic images. These synthetic images feature two important challenges that can only be resolved by weighted curvature regularization: 1) Absence of data differences at some parts of the boundary, \textit{i.e.}, object and background have the same intensity profile. 2) Segmentation of high curvature features such as cusps and sharp corners.

Figure 2 shows three images that the human visual perception can seamlessly define where the boundaries of objects of interests are. However, some of the most advanced segmentation tools can not. For example, in the first image of Figure 2, it is obvious that the blue seed should provide an object that ends where the white lines end. Graph cuts favor the cut with the minimal number of edges since there no contrast between the neighboring pixels which yields a trivial solution isolating the blue seed as a foreground and the rest of the image as a background, graph cuts also gave trivial solutions in the rest of the images. Random walker works, intuitively, by calculating the probability that a random walk starting at a particular pixel will first reach one of the seeds. Hence, it suffers from a proximity problem that results in a premature stopping and segments a bar with a smaller length than the correct one.  Our approach provides the correct segmentation. Although the seed is very far from the end of the bar,  our algorithm could extend the segmentation until the end of the bar. This is simply because a a straight line has a minimal curvature so the algorithm extends the bar until an intensity difference occurs.

\begin{figure}[h!]
\vspace{-0.5in}
\begin{center}
  \includegraphics[width=0.9\linewidth]{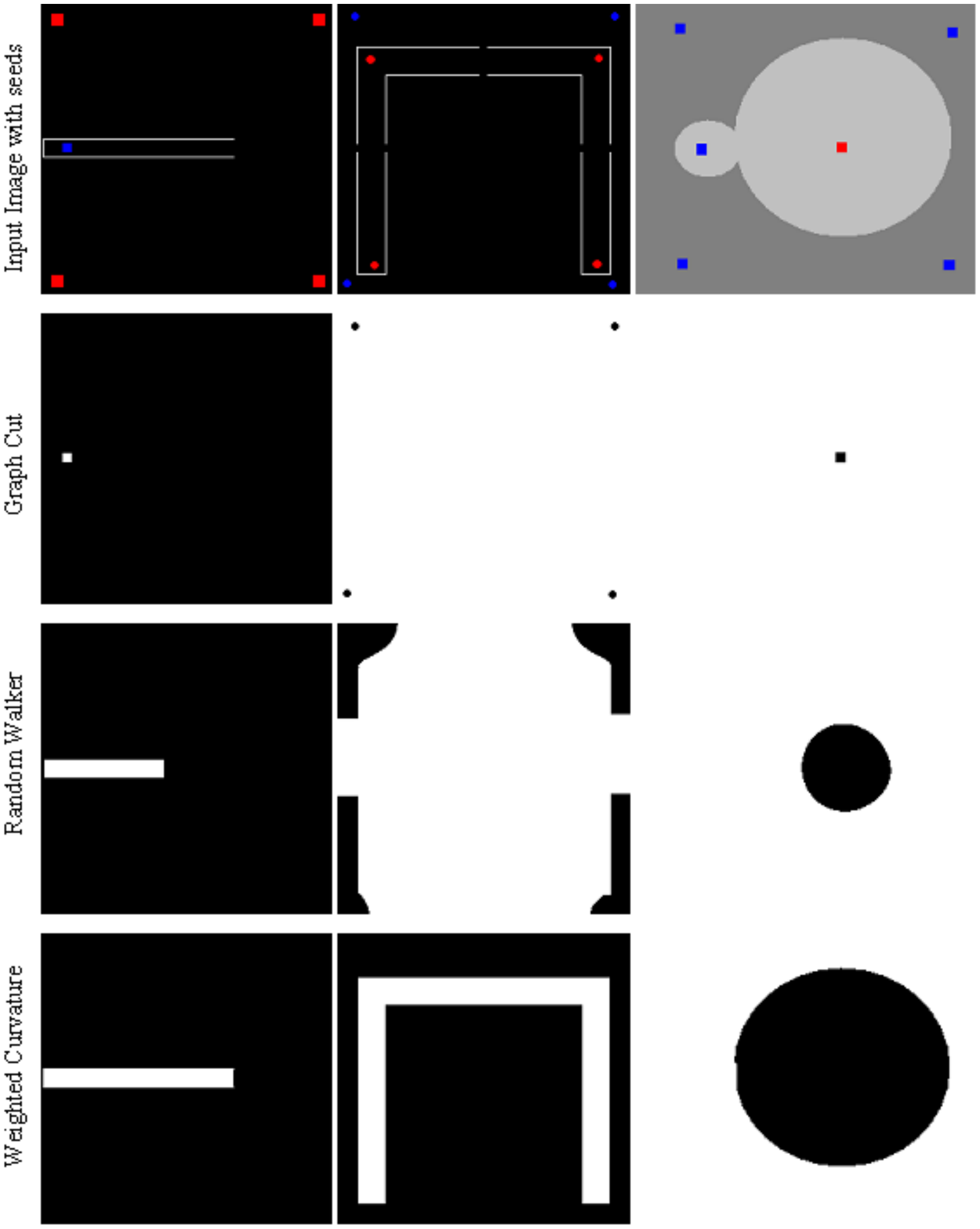}
  \vspace{-0.3in}
  \caption{Segmentation of some challenging images. Graph cuts provide a trivial solution by isolating the foreground or the background seeds. Random walker suffers from proximity problem that yields a smaller object than the desired in images 1 and 3. It also suffers from a leakage producing a segmentation of the second image due to the gaps in the boundary. Weighted curvature extends the bar in  the first image at no cost and stops when contrast change occurs. It bridges the gaps in the second image because curvature preserves connectivity. In the third image, the curvature regularization separates two structures of the same intensity.}
    \end{center}
\label{Challenge1}
\vspace{-0.3in}
\end{figure}

\begin{figure}[h!]
\vspace{-1in}
\begin{center}
  \includegraphics[width=0.9\linewidth]{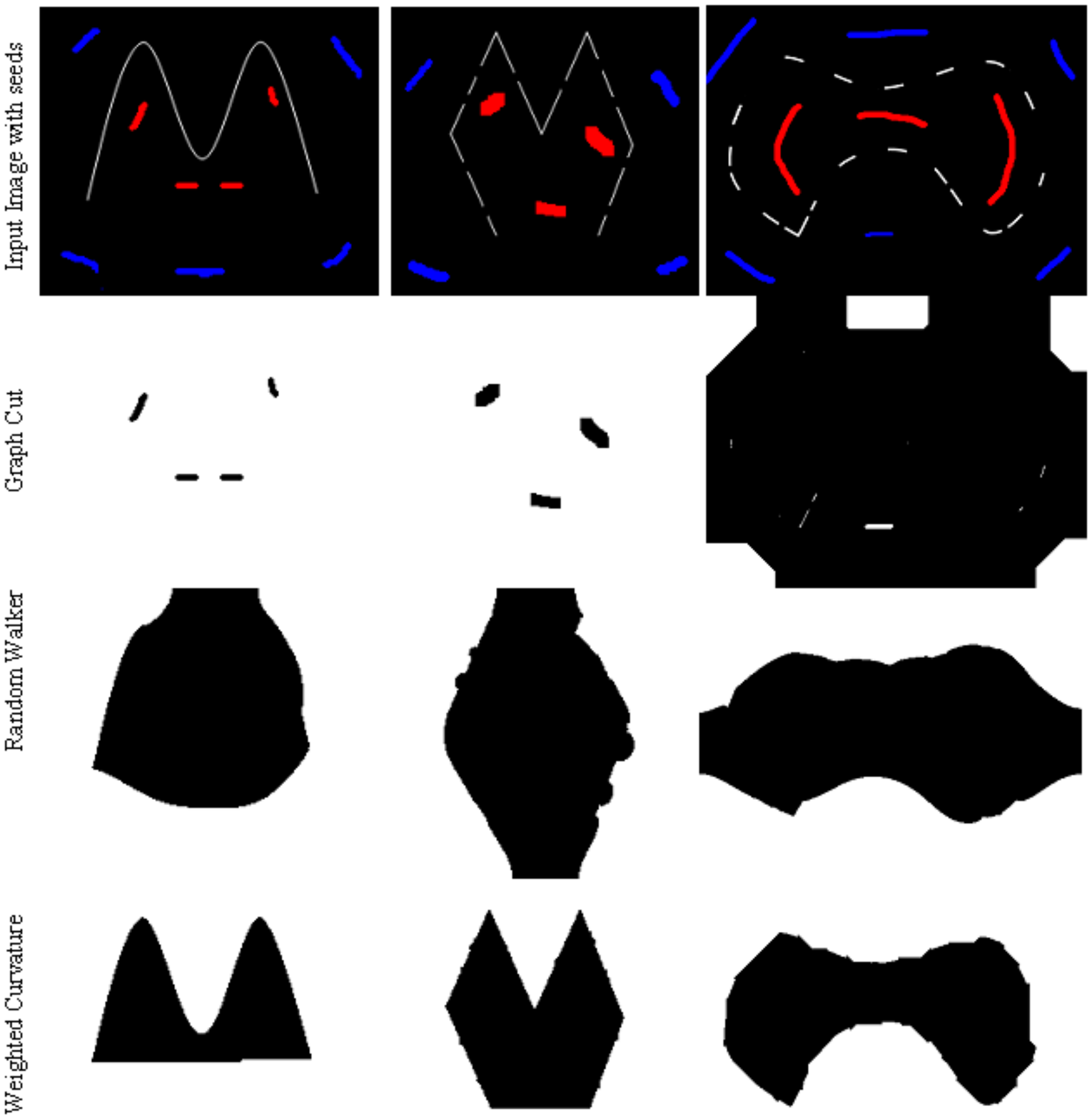}
  \vspace{-0.8in}
  \caption{Segmentation of boundaries with very large gaps and zero contrast between foreground/background pixels. The gaps caused leakage in the random walker that failed to provide the desired segmentation. Curve continuity is promoted by curvature regularization that enables our approach to  bridge the gaps in the boundaries of the three images. Our approach also maintains the cusps in the first two images because they are warranted by the contrast weighted edges. }
    \end{center}
\label{Challenge2}
\vspace{-0.3in}
\end{figure}
The second image features a disconnected boundary. This challenge is common in real images when acquisition artifacts such as noise and occlusion disconnect the boundary. In medical images, anatomical abnormalities such as stenosis in vessels may cause the boundary to appear disconnected.  The random walker leaks through these gaps and fails to separate the object from the background correctly. Weighted curvature, however, succeeds to bridge the gaps in the boundary due to the unique ability of curvature to preserve object continuity as connected boundaries have less curvature than disconnected ones.  Meanwhile, the sharp corners were not smeared by curvature minimization because they were supported by contrast and the curvature is weighted by this contrast information in our formulation. The third image consists circle with a bump (of the same intensity) attached to it. The large circle is the object of interest that should be separated from the smaller bump and the black background. This scenario is very common in medical images. For example, a tumor may be made of the same tissue as the organ it is attached to, with no intensity differences between the tissue and the tumor (see Figure 6) or two proximal distinct structures may have the same intensity profile (such as the caudate and the putamen in brain MRI, for example). In figure 2, random walker stops prematurely yielding a smaller circle than the correct one because a random walk from an erroneously-background-labeled pixel would have a higher probability reaching the background seeds than the foreground one. However, weighted curvature succeeds to obtain the correct boundary; the contrast based weights forces the boundary to stop when it hits a black pixel isolating the large circle from the background. And the curvature favors a circle with larger radius and hence it extends the foreground seeds until it reach the boundary between the large circle and the smaller bump. The foreground can not be extended any further, otherwise, it would form a cusp at the blue seed in the bump yielding a high curvature. Premature segmentation is prohibited by the fact that a circle with a larger radius has a smaller curvature thana circle with a smaller radius. Figure 3 shows similar examples with larger gaps in the boundaries. Notice that although the gaps in the dotted curve (in the first image) are very small, these gaps caused the random walker to leak. Also, random walker could not connect the large gap, between two end points of the curve, by a straight line. Weighted curvature, however, prefers the straight line because of its minimal curvature which wields the desired segmentation.
The second and third images in Figure 3 also feature very large gaps in the boundaries with no foreground/background contrast, these gaps challenged the random walker resulting in erroneous segmentation. Weighted curvature succeeds to bridge all the gaps in the boundaries. Notice that the cusps in the first image and the sharp corners in the second image were segmented correctly because the curvature weights are contrast dependent.

\begin{figure}
\vspace{-0.1in}
\begin{center}
\includegraphics[width=1\linewidth]{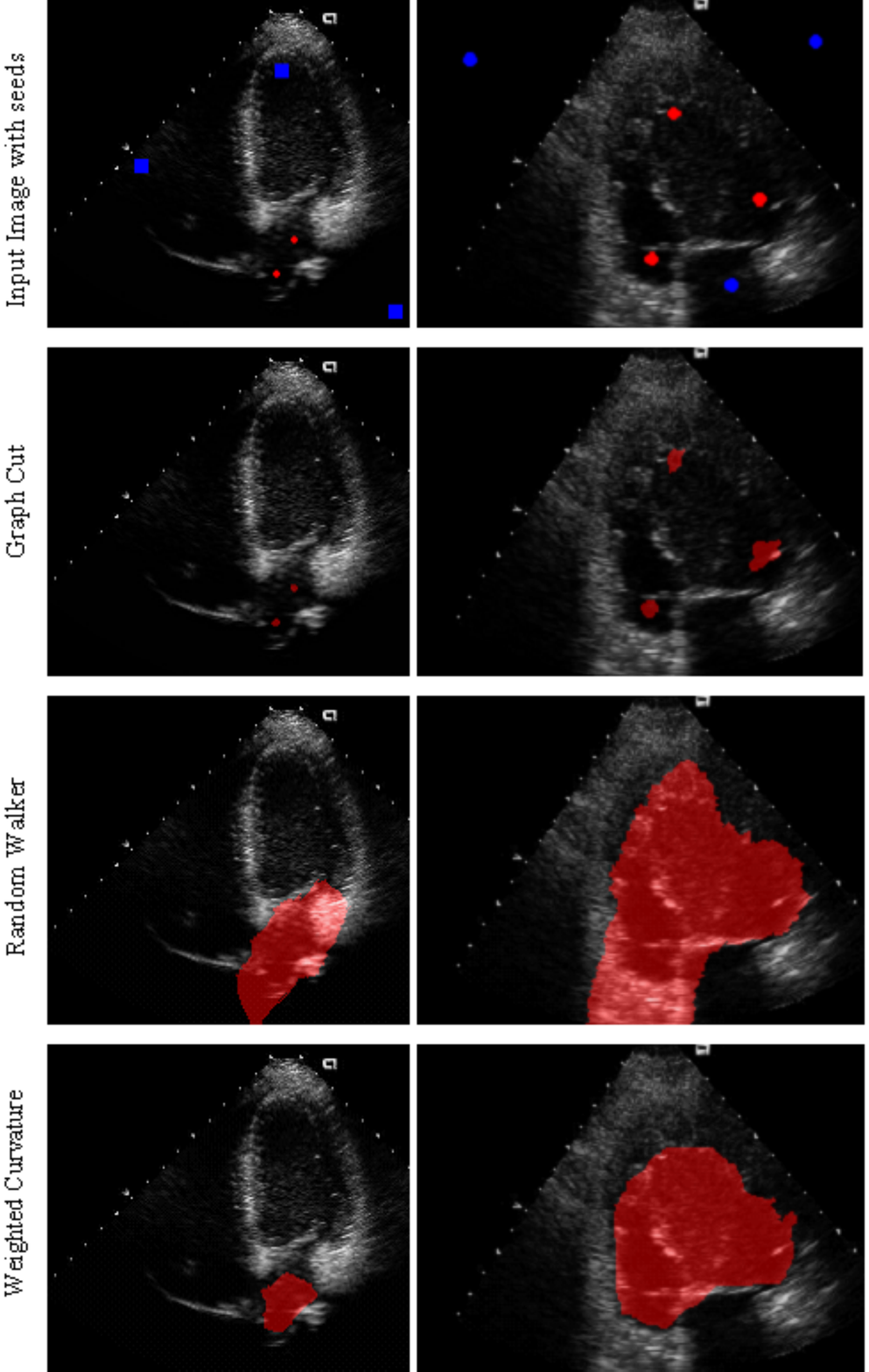}
\vspace{-0.1in}
\caption{First column: Segmentation of the left Atrium. Second Column: Segmentation of the left ventricle. Graph cuts yielded a trivial segmentation around the seeds. The lack of contrast between the left atrium and the background, in addition to the proximity of several parts of the images to the foreground seeds forced the random walker to leak and produce an overcompensation of the region of interest in both cases. Weighted curvature prevented leakage because leakage introduces higher curvature cost of the boundary.}
\end{center}
\vspace{-0.25in}
\label{Ultrasound}
\end{figure}
\begin{figure}
\vspace{-0.1in}
\begin{center}
  \includegraphics[width=1\linewidth]{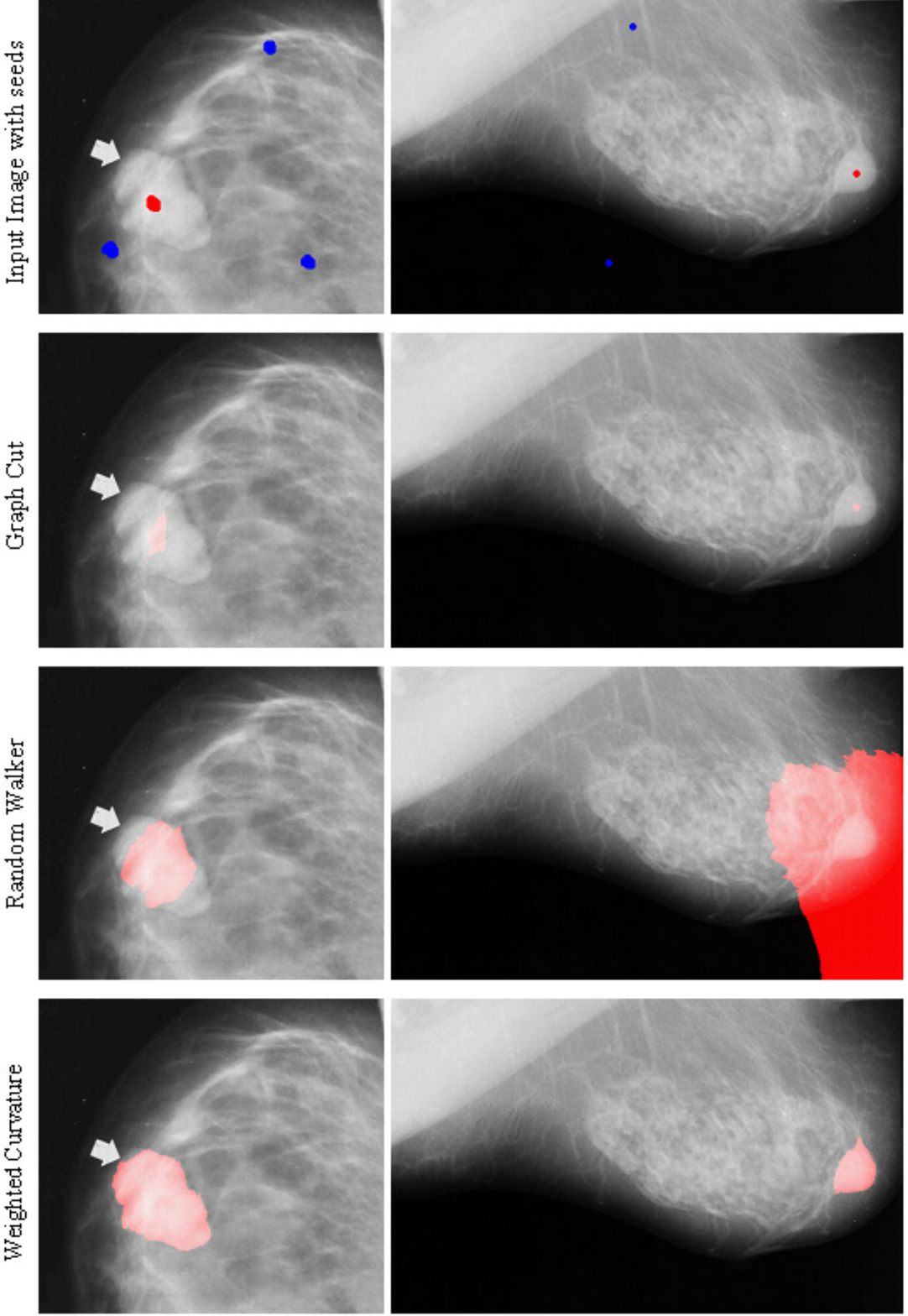}
  \vspace{-0.25in}
  \label{Mammogram}
  \caption{Segmentation of circumscribed masses in mammograms. Graph cuts resulted in a trivial solution by cutting the edges around the foreground seeds. Random walker resulted in false negatives in the segmentation of the first mass and false positives in the segmentation of the second one because the probability of a random walk reaching a foreground or background seed is affected by the location of the seeds yielding a false segmentation in both cases. Weighted curvature detected the correct boundaries in both cases. }
\end{center}
    \vspace{-0.25in}
    \end{figure}

    \begin{figure}
\vspace{-0.1in}
\begin{center}
  \includegraphics[width=1\linewidth]{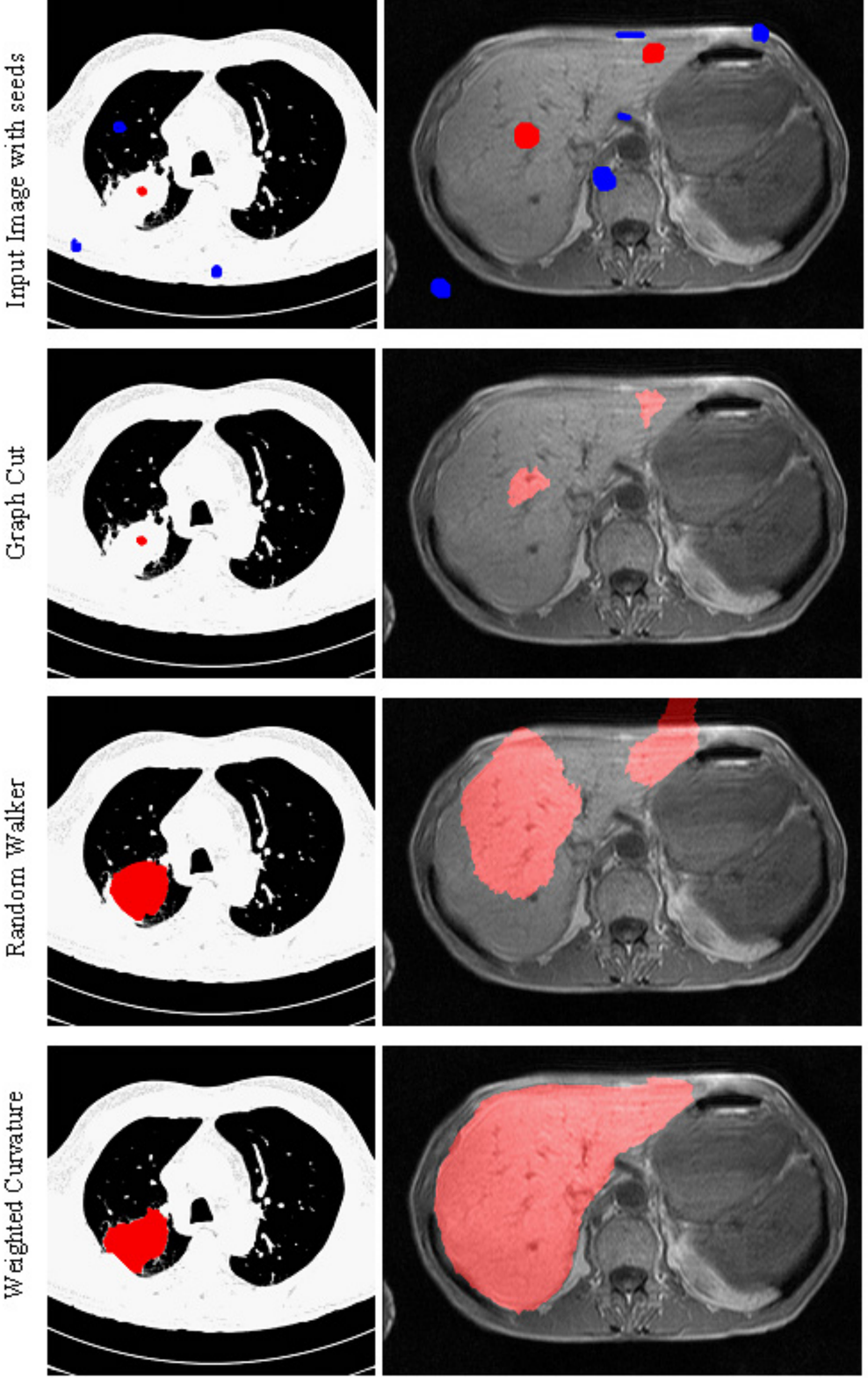}
  \vspace{-0.25in}
  \label{LungLiver}
  \caption{First column: Segmentation of a lung tumor in a chest CT scan. Graph cuts produced a trivial segmentation around the foreground seed. Random walker did not extract the whole tumor due to the proximity of some tumor pixels to background seeds. Weighted curvature correctly classified the pixels that random walker missed due ti their high contrast with the lung tissue.  Second column: segmentation of the liver in an abdominal MRI slice. Random walker yielded false positives and false negatives and disconnected the liver into two pieces while weighted curvature preserved the connectivity of the liver and captured the correct boundary. }
\end{center}
    \vspace{-0.25in}
    \end{figure}

The previous challenges occur frequently in medical images. Here we demonstrate the performance of our model on several medical imaging applications and different modalities. Figure 4 shows the segmentation of the left atrium in ultrasound. The atrium has the same intensity profile as the rest of the heart chambers and the background with no clear boundaries. Graph cuts yielded a trivial solution isolating the foreground seeds as the object and the rest of the image as a background. The absence of image contrast between the atrium and the background resulted in an over-segmentation by random walker. The segmentation using our proposed approach succeeded to extract the left atrium correctly.
The second columns of Figure4 exhibits another example for such challenges where the left ventricle is the object to be segmented.
Mass segmentation in mammograms is a challenging problem that can not easily be modeled by data terms because of the large variability among the different types of breast tissues (e.g fatty, glandular and dense tissues) and because of the variability in the appearance and shape of the masses . Figure 5 demonstrates the segmentation of two circumscribed masses in mammograms. For the first mass, the random walker result suffers from false negatives (under-segmentation) while the second mass the random walker provided  false positive pixels (over-segmentation). In both cases, the weighted curvature yielded a correct segmentation.
Figure 6 depicts the segmentation of a lung tumor in a chest CT scan. There is no contrast between the tumor and abdominal muscles.  Random walker stopped prematurely (due to the proximity issue) and did not extract the whole tumor while our segmentation did. The weighted curvature expands the foreground minimize the curvature of its boundary and it stops when it reaches the boundary between the tumor and the muscles that produces the least curvature. The second column of figure 6 shows a similar case where there is no contrast between the liver and the abdominal muscles in an abdominal MR scan. Random walker yielded some false positives and and false negatives. For example, the upper boundary of the liver was disconnected by random walker. It is intuitively clear that random walks starting at these erroneously-background-labeled pixels (the gap between the two pink parts captured by random walker) have higher probability to reach background seeds than foreground and hence falsely labeled as background. However, curvature regularization favors a smooth boundary over a disconnected one because disconnecting the upper boundary into two pieces would form corners with high curvature.
All the previous results were obtained using the same parameters for attraction force and contrast weighting. Complete global solution was obtained in each case. The addition of the attraction force enabled the QPBO to provide complete labeling without any need to perform probing which makes our algorithm even faster than \cite{CurvatureCVPR10}. In the few cases that QPBO did not label all the pixels, the probing was able to provide these pixels yielding a global optimal solution.

\section{Conclusion}

We proposed an image segmentation method which isolates an object
boundary by using weighted curvature. Weighted curvature allows us to
segment an object in challenging situations in which a data term is
unreliable or missing, as is common in medical image segmentation
tasks.  Additionally, our formulation of weighted curvature permits
the segmented object boundary to exhibit a high curvature if it is
warranted by the image data.  The inclusion of an attraction force
allows us to find a global optimum of the model using QPBOP in a few
seconds.

Our experiments demonstrate that our model can be used to complete
object boundaries in controlled experiments where the
object/background shared the same intensity distribution and
significant parts of the boundary were missing.  In contrast, other
leading algorithms which are known for this type of robust behavior
were unable to achieve segmentations of these challenging control
images.  Similarly, our algorithm was demonstrated to provide quality
segmentations on image segmentation tasks appearing in medical images
which exhibit similar challenges to the control images. Once again, we
demonstrate that graph cuts and random walker are unable to perform
with such little information (in terms of image data and seed
numbers/placements). In all of these experiments, both control and
real, the \emph{same} parameters were used for the contrast weighting
and attraction force term.

Future work will be to investigate the use of other optimization
methods that might allow us to remove the attraction force (and
consequent parameter), the use of a data term in conjunction with
weighted curvature and to take advantages of recent advances in
GPU-based graph cuts (and therefore QBPOP) to further accelerate the
already fast speed of our method.

\bibliographystyle{splncs}
\bibliography{ECCV_bib}

\end{document}